\pdfoutput=1

\documentclass[11pt]{article}

\usepackage{acl}

\usepackage{times}
\usepackage{latexsym}

\usepackage[T1]{fontenc}

\usepackage[utf8]{inputenc}

\usepackage{microtype}

\usepackage{import}
\usepackage{fancyvrb}
\usepackage{xspace} 
\usepackage{makecell}
\usepackage{amssymb}
\usepackage[colorinlistoftodos]{todonotes}

\newcommand{\vtick}{\textquotesingle}

\newcommand{\ppceme}{PPCEME\xspace}
\newcommand{\ppcmbe}{PPCMBE\xspace}
\newcommand{\ppcme}{PPCME2\xspace}
\newcommand{\ptb}{PTB\xspace}
\newcommand{\eebo}{EEBO\xspace}

\newcommand{\pos}{POS\xspace}
\newcommand{\evalb}{evalb\xspace}
\newcommand{\elmo}{ELMo\xspace}
\newcommand{\bert}{BERT\xspace}

\newcommand{\lstm}{LSTM\xspace}
\newcommand{\ftagsa}{ftags-0\xspace}
\newcommand{\ftagsb}{ftags-10\xspace}
\newcommand{\ftagsc}{ftags-31\xspace}

\title{Penn-Helsinki Parsed Corpus of Early Modern English: First Parsing Results and Analysis}

\author{Seth Kulick \and Neville Ryant \\
Linguistic Data Consortium \\
University of Pennsylvania \\
\texttt{\{skulick,nryant\}@ldc.upenn.edu} \\
\And 
Beatrice Santorini \\
Linguistics Dept. \\
University of Pennsylvania \\
\texttt{beatrice@sas.upenn.edu} \\
}

\begin{document}
\maketitle
\begin{abstract}

We present the first parsing results on the Penn-Helsinki Parsed Corpus of Early Modern English (PPCEME), a 1.9 million word treebank that is an important resource for research in syntactic change. We describe key features of PPCEME that make it challenging for parsing, including a larger and more varied set of function tags than in the Penn Treebank. We present results for this corpus using a modified version of the Berkeley Neural Parser and the approach to function tag recovery of \citet{gabbard-etal-2006-fully},
Despite its simplicity, this approach works surprisingly well, suggesting it is possible to recover the original structure with sufficient accuracy to support linguistic applications (e.g., searching for syntactic structures of interest). However, for a subset of function tags (e.g., the tag indicating direct speech), additional work is needed, and we discuss some further limits of this approach.  The resulting parser will be used to parse Early English Books Online, a 1.1 billion word corpus whose utility for the study of syntactic change will be greatly increased with the addition of accurate parse trees.
 

\end{abstract}


\section{Introduction}
\label{sec:intro}

The Penn-Helsinki Parsed Corpus of Early Modern Eng\-lish (\ppceme) \citep{ppceme} consists of over 1.7
million words of text from 1500 to 1712, manually annotated for phrase structure.  It belongs to a family of treebanks of historical Eng\-lish \citep{ycoe, ppcme2, pceec, ppcmbe} 
and other languages \citep{icepahc, tycho-brahe, mcvf, ppchf} with a shared annotation philosophy and similar guidelines across languages, which form the basis for reproducible studies of syntactic change \citep{kroch-taylor-ringe, ecay, wallenberg-2016, galves-20, wallenberg-et-al-2021}.


Properties of \ppceme make it of interest for NLP researchers as well as linguists. The source text lacks the standardization of spelling and punctuation present in treebanks such as the  Penn Treebank (\ptb) \citep{marcus-etal-1993-building}, and while the annotation style resembles that of the \ptb it features some important differences.  These include more finely-grained Part of Speech (\pos) tags, differences in how particular structures are annotated, and perhaps most significantly, the increased use of function tags.  There has however been relatively little work\footnote{Google scholar only revealed three papers using these corpora: \citet{kulick-etal-2014-penn}, which parses \ppcmbe; \citet{moon-baldridge-2007-part}, which explores \pos-tagging for \ppceme; and \citet{yang-eisenstein-2016-part}, which explores \pos-tagging for \ppceme and \ppcmbe.} in the NLP community using \ppceme or its sister corpora, the Penn Parsed Corpus of Middle English, 2nd edition (\ppcme) and the Penn Parsed Corpus of Modern British English (\ppcmbe).\footnote{All three corpora are collected in \citet{ppche}.}

In this paper, we establish baseline results for both the parsing accuracy and function tag recovery for \ppceme, pointing out some limitations for the latter.
Our parser is a slightly modified version of the self-attentive neural constituency parser introduced by \citet{kitaev-klein-2018-constituency} with function tags recovered using the approach of \citet{gabbard-etal-2006-fully}. Our main contributions are: (1) the first parsing results on \ppceme, 
(2) demonstration of the successes but also limitations of the function tag recovery, and (3) further evidence of the importance of in-domain pretraining for contextualized word representations.

The larger context of this work informs some of our decisions in what follows.  While \ppceme is relatively large for a manually annotated corpora, its usefulness for linguistic research is limited by the fact that various phenomena of interest are too sparse within it to support reliable statistical models of language change.  In contrast, the Early Eng\-lish Books Online (\eebo) corpus \citep{eebo} covers the same time period,  1475 to 1700, with  1.5 billion words of text.  However, \eebo's potential as a resource for linguistic research remains unrealized because it is not linguistically annotated and its size renders manual annotation infeasible.   The parsing model trained on \ppceme will be used to parse \eebo for this purpose. Therefore, while the \ppceme work described here is self-contained, our focus on particular aspects (function tags, some \pos tags) is due to consideration of this larger longer-term goal.

\section{Differences between PPCEME and PTB}
\label{sec:ppceme}
As mentioned in the introduction, the \ppceme annotation style differs from that of \ptb in certain respects.  We describe here those differences in more detail, and how in some cases they  impacted on our preprocessing.

\subsection{\ppceme Part-of-Speech Tags}
\label{sec:ppceme:pos}

\subsubsection{Complex Tags}

Although we generally attempt to avoid modifying the existing annotation, \ppceme's very large set of POS tags (N = 353) requires trimming to a computationally more tractable size. 

Of the 353 tags just mentioned,  213 are complex tags intended to facilitate tracking changes in orthographic conventions over time - for instance, the development of {\tt (ADJ gentle) (NS men)} to {\tt (ADJ+NS gentlemen)}.  Since these changes are irrelevant for present purposes, we prune such tags in accordance with the Righthand Head Rule, yielding {\tt (NS gentlemen)}.\footnote{\citet{yang-eisenstein-2016-part} simplify the complex tags for the same reason as we do, but keep the leftmost tag, which for English is incorrect in the general case.}   Certain rare cases, such as {\tt (WPRO+ADV+ADV whatsoever)} or {\tt (Q+BEP+PRO albeit)}, are exceptions to the Righthand Head Rule.  In such cases, the best simple tag is sometimes the leftmost tag and sometimes another tag entirely ({\tt (WPRO whatsoever), (P albeit)}).  We simply ignore this complication on the grounds that these cases are a small subset of the complex tags, which themselves are used for only about 1\% of the words in the corpus.  After pruning and some other minor changes discussed in Appendix \ref{app:ppcemepre}, 85 POS tags remain.


\subsubsection{Distinctions among Verb Classes}

\ptb makes no distinction between main verbs and the auxiliary verbs {\it be}, {\it do} and {\it have}, but this distinction is vital for us, since it is exactly the syntax of main (but not auxiliary) verbs that changes 
over the course of Early Modern English.  In fact, even among the verbs with auxiliary uses, we need to distinguish {\it do} from the other auxiliaries in order to track the rise of auxiliary {\it do}. For this reason, we do not follow \citet{yang-eisenstein-2016-part} in mapping the \ppceme tags for verbs to the smaller set used in \ptb.

\subsection{\ppceme Phrase Structure}
\label{sec:ppceme:ps}



\subsubsection{Function Tags}
\label{sec:ppceme:ps:ftags}

In phrase-structure treebanks, function tags can be appended to syntactic category labels in order to provide information about a constituent's grammatical or semantic role.   The \ptb uses 20 function tags in this way, while distinguishing other constituent roles by means of structural differences (e.g., adjoining a relative clause but not a complement clause)
By contrast, \ppceme relies on function tags uniformly, largely because it has neither base NPs or VPs.  As a result, \ppceme's set of function tags is larger than \ptb's. Omitting a few rare types, we consider 31 function tags in the work reported below.\footnote{See Appendix
\ref{app:ppcemepre} for the details, along with some information on  function tag frequency.}  
The following tree illustrates \ppceme's use of function tags to encode central grammatical roles. The subject and indirect object are sisters, but distinguished by the function tags SBJ and OB2, respectively. MAT and SUB on the two IPs identify the higher one as a matrix clause and the lower one as a subordinate clause. Finally, THT indicates that the CP is a {\it that} complement clause (rather than, say, a relative or adverbial clause). 

{\vbox
{\footnotesize
\begin{verbatim}
(IP-MAT (CONJ and)
        (NP-SBJ (D the) (N schereffe))
        (VBD shewed)
        (NP-OB2 (PRO$ my) (N servant))
        (CP-THT (C that)
                (IP-SUB ...)))
\end{verbatim}
}
}


There has been some work on recovering function tags in \ptb \citep{blaheta-charniak-2000-assigning,blahetadiss,gabbard-etal-2006-fully,merlo-musillo-2005-accurate}, but overall they have received only limited attention. We are not aware of any work to recover the function tags in the historical corpora.  Given the centrality of certain function tags (notably, SBJ) for retrieving the sentence types of interest to us, we are constrained to include them in the parsing model.  

We use three different versions of the training and evaluation files discussed in the next section, differing only in which function tags are included.  One version (``\ftagsa'') includes no function tags at all (e.g. {\tt NP-SBJ} becomes {\tt NP}), for the second  we used ten of the tags that were of particular importance for some of the linguistic searches (``\ftagsb''), and  for the third all 31 tags were used (``\ftagsc''). We give a detailed description of the tags in the context of the parser analysis in Section \ref{sec:results:ftags}.

\section{Cross-validation Splits}
%
\label{sec:split}

Recently, concerns have been raised over the validity of inferences drawn from static train/dev/test splits of a corpus; for instance, see  \citet{gorman2019we}, who evaluate the consistency of rankings of POS taggers across 20 random splits of the WSJ section of \ptb. For us, this issue is particularly pressing as \ppceme contains relatively few individual source texts, thus increasing the chance that a single particularly difficult or non-representative source text will greatly skew performance on the dev/test partitions.  

We therefore define an 8-fold cross-validation split, with each component split roughly matching the
90\%-5\%-5\% distribution in the standard single \ptb split.  Within each partition (train, dev, test) of a split, we attempted to equally represent (in terms of equal word counts) each of \ppceme's three time periods, as indicated  by ``e1'', ``e2'', and ``e3'' in the filenames.  Given our eventual goal of parsing all of \eebo, which encompasses all of these time periods, this step is necessary in order to adequately predict performance on that corpus.\footnote{By contrast, \citet{yang-eisenstein-2016-part}, split \ppceme into thirds by time period (rather than across time periods) for the different purpose of studying domain adaptation.}  Finally, in cases where \ppceme distributes a single source text over several annotated files, we were careful to assign all such files to the same partition.  As \ppceme contains 448 annotated files, but only 232 distinct source texts, this greatly constrained how we could define the partitions.  Nevertheless, we succeeded in including 209 of the 232 source texts in either a dev or test partition of one of the 8 splits.  For more details on the split definitions, see Appendix~\ref{app:splits}.

\section{ELMo Embeddings Trained on \eebo}
\label{sec:elmoeebo}
In recent years, contextualized word embeddings such as \elmo \citep{peters2018deep} and  \bert \citep{devlin-etal-2019-bert} have driven significant improvements on downstream NLP tasks, including \pos tagging and parsing. Due to the significant overhead involved in training these representations, researchers often use pretrained models distributed by large companies, sometimes fine-tuned to the domain of interest.  Although this often produces perfectly satisfactory results, in cases of significant mismatch between a test domain and standard training domains - usually sources such as text scraped from Wikipedia, BooksCorpus \citep{zhu2015aligning}, and news text from Common Crawl \citep{nagel}, and discussion forums \citep{radford2019language} - pretraining on the novel domain yields significant improvements \citep{lee2019biobert,beltagy2019scibert,jin2019probing}.

Because of the orthographic and syntactic differences between Early Modern Eng\-lish and contemporary Eng\-lish mentioned in Section \ref{sec:ppceme},  our current work involves exactly such a mismatch, and so we pretrained \elmo embeddings on \eebo.\footnote{At present, we lack  the computational resources for the obvious next step of pretraining \bert embeddings on \eebo, but we are pursuing access to them.} We used the same model configuration as \citet{peters2018deep} for 11 epochs\footnote{This corresponds to 2 weeks of training using 4 GTX 1080 GPUs.} using all of 1.1 billion words of \eebo. Pretrained was performed using the TensorFlow implementation maintained by AllenNLP\footnote{\url{https://github.com/allenai/bilm-tf}} using the default model configuration. 
We then integrated the resulting embeddings, which have 1,024 dimensions, into the parser model, as discussed in Section \ref{sec:model}. 

See Appendix \ref{app:elmoeebo} for further details regarding the materials used for pretraining.

\section{Model and evaluation}
\label{sec:model}
\subsection{Parser Architecture}
\label{sec:model:parser}
We use the parsing model of \citet{kitaev-etal-2019-multilingual}, which represents a  constituency tree $T$ as a set of labeled spans $(i, j, l)$ where $i$ and $j$ are the beginning and ending positions of the span and $l$ its label. Each tree is assigned a score $s(T)$, which is decomposed as a sum of per-span scores:
    \begin{equation}
        s(T) = \sum_{(i,j,l) \in T} s(i,j,l)
    \end{equation}
The per-span scores $s(i,j,l)$ are themselves assigned using a neural network that takes a sequence of per-word embeddings as input, processes these embeddings using a transformer-based encoder \citep{vaswani2017attention}, and then produces a span score from an MLP classifier \citep{stern-etal-2017-minimal}. The highest scoring valid tree is found using a variant of the CKY algorithm. \pos tags are recovered using a separate classifier operating on top of the encoder output, which is jointly optimized with the span classifier. For more details, see \citet{kitaev-klein-2018-constituency}.

We consider four approaches for producing the word embeddings that serve as input to the encoder: 

\noindent \textbf {bert-base} Contextual word embeddings are computed using \bert \citep{devlin-etal-2019-bert} by retaining the output of the last layer for the final sub-word unit of each token. We use the {\it bert-base} model distributed via the {\it transformers} Python package\footnote{\url{https://github.com/huggingface/transformers}}, which is pre-trained on 3.3 billion words of modern Eng\-lish.

\noindent \textbf {elmo (orig)} As above, but using \elmo \citep{peters2018deep}. We use the \elmo-original model distributed by AllenNLP\footnote{\url{https://allennlp.org/elmo}}, which was pre-trained on one billion words of modern Eng\-lish.

\noindent \textbf {elmo (eebo)} \elmo embeddings pre-trained on 1.1 billion words of early modern Eng\-lish from \eebo. Uses the same model architecture as {\it elmo (orig)}. See also Section~\ref{sec:elmoeebo}.

\noindent \textbf{char-lstm} Tokens are represented as the concatenation of the outputs of a bidirectional character-level \lstm for the final character of the token.

Our implementation is based on  version 0.2.0 of the Berkeley Neural Parser\footnote{\url{https://github.com/nikitakit/self-attentive-parser}} with modifications to allow the use of \elmo. 
We train each of the 8 models (one for each cross-validation split) for 50 epochs, using the evalb score on the dev section as our criterion for saving  the best model. For Additional details regarding training and hyperparameters, see Appendix~\ref{app:model:parser}.

\subsection{Function Tags}
\label{sec:model:ftags}

We adopt the approach of \citet{gabbard-etal-2006-fully} for function tag recovery, in which the parser preprocessing step simply does not delete the function tags, and so nonterminals such as {\tt NP-SBJ} are treated as an atomic unit. Since the decision whether to delete is part of the preprocessing, this approach does not require modification to the parser. As mentioned in Section 
\ref{sec:ppceme:ps:ftags}, our preprocessing code for \ppceme creates three versions of the files, \ftagsa, \ftagsb, and \ftagsc.

\subsection{Parsing and Function Tag evaluation}
\label{sec:model:evaluation}
During training, the parser uses the standard  \evalb \citep{evalb} on the dev set to determine the  best model so far. However, since our models also predict \pos tags, and  \evalb removes punctuation based on \pos tags, inconsistent sentence lengths can arise if the gold and parsed trees have differing \pos tags, resulting in ``Error'' sentences in the \evalb output.  For the scores reported in Section \ref{sec:results:parser}, we therefore use the modified \evalb supplied with the Berkeley parser, originating from \citet{seddah-etal-2014-introducing}, which does not delete any words, so the differences in \pos tags do not affect the sentence length.

Function tags are typically removed by \evalb for the comparison of bracket labels,  and we have not modified this.  To evaluate the function tag recovery, we follow the approach of \citet{gabbard-etal-2006-fully} which in turn follows \citet{blahetadiss}.  This evaluation compares function tags only for nonterminals that are counted as matches for \evalb,   For example,  if  a NP-SBJ node in the gold tree matches with a NP-SBJ node in the parsed tree,  it is a match for SBJ, while if a NP-SBJ node in the gold tree matches with a NP-OB1 node in the parsed tree (which can happen since the function tags do not count for \evalb), it is a recall error for SBJ and a precision error for OB1.

\begin{table}[t]
\footnotesize
    \centering
\begin{tabular}{|c|c|c|c|} \hline
config   & ftags-0        & ftags-10       & ftags-31 \\ \hline
\multicolumn{4}{|c|}{{\it dev}} \\  \hline
bert     & 90.37 (1.95) & 90.37 (1.95) & 90.37 (1.98) \\ \hline
elmo-e   & 90.92 (1.86) & 90.95 (1.79) & 90.89 (1.83) \\ \hline
elmo-o   & 88.06 (2.23) & 88.13 (2.25) & 88.09 (2.22) \\ \hline
char     & 87.28 (2.34) & 87.43 (2.33) & 87.33 (2.33) \\ \hline
\multicolumn{4}{|c|}{{\it test}} \\  \hline
bert     & 89.81 (0.76) & 89.83 (0.80) & 89.86 (0.77) \\ \hline
elmo-e   & 90.56 (0.66) & 90.62 (0.71) & 90.53 (0.69) \\ \hline
elmo-o   & 87.31 (0.85) & 87.50 (0.84) & 87.38 (0.81) \\ \hline
char     & 86.29 (0.93) & 86.50 (0.96) & 86.45 (0.90) \\ \hline
\end{tabular}
    \caption{Cross-validated F1 for the parser on \ppceme using no function tags, 10 function tags, or 31 function tags. The standard deviation for F1 is presented in parentheses. Scores are obtained using the version of \evalb that does not delete punctuation, for reasons discussed in Section~\ref{sec:model:evaluation}, and do not consider function tags in the matching of brackets. {\it bert}: bert-base, {\it elmo-e}: \elmo pretrained on \eebo, {\it elmo-o}: original \elmo embeddings, {\it char}: character \lstm}
    \label{tab:cvscores}
\end{table}

\section{Pretraining comparison experiments}
\label{sec:results:parser}
Table~\ref{tab:cvscores} presents parsing results for the dev/test sections  of the 8 cross-validation splits described in Section~\ref{sec:split}.  The rows are the four embedding representations described in  Section~\ref{sec:model:parser} and the columns are the F1 scores (evalb bracket scores, as discussed in Section \ref{sec:model:evaluation}) for each of the three versions  (\ftagsa, \ftagsb, \ftagsc) of the training and evaluation data.  Each cell shows the mean and standard deviation for that embedding representation and function tag version over the 8 splits.  As there is no great precision/recall imbalance, we relegate the corresponding precision and recall numbers to a more complete  version of this table in Appendix~\ref{app:pretraining}.

For each of the three versions of the training and evaluation data, the best F1 scores are obtained using \elmo (\eebo),  followed by \bert, \elmo (orig), and, char-\lstm. Moreover, \elmo (\eebo) outperforms \bert (to varying degrees) on every single combination of split/function tags. In particular, for the version with 31 function tags, which will be the focus of further analysis in the following sections, \elmo (\eebo) outperforms \bert by 0.67\% absolute (t=6.05, p$<$1e-3) on the test set. 

The fact that \bert is consistently outperformed by  \elmo (\eebo) is particularly interesting given that (a) \bert in general outperforms \elmo when trained on similar material, and (b) \bert in this case was trained on three times as much material as \elmo. This result underlines the importance of pre-training on in-domain materials, a fact that has also been observed for NLP tasks in biomedical \citep{lee2019biobert}, financial \citep{araci2019finbert}, and legal \citep{chalkidis2020legal} domains.


Overall, we note that the scores are a few points lower than the current state of the art
for \ptb (at submission time, 96.38\% F1 on the test section \citep{mrini2019rethinking}).
As \citet{kulick-etal-2014-penn} point out, all of the English historical corpora lack certain brackets present in \ptb (base-NPs and VPs) that are relatively ``easy to get'', and this tends to adversely affect their parsing scores.  Specifically,  \citet{kulick-etal-2014-penn} find the F1 score for \ppcmbe to be lower than for \ptb by about 2\% absolute, an effect we expect carries over to \ppceme.

\section{Part-of-speech Results}
\label{sec:posresults}

\begin{table}[t]
\footnotesize
    \centering
\begin{tabular}{|c|c|c|c|} \hline
  config  & ftags-0      & ftags-10     & ftags-31 \\ \hline
  \multicolumn{4}{|c|}{{\it dev}} \\  \hline
  bert    & 97.75 (0.78) & 97.73 (0.83) & 97.71 (0.89) \\ \hline
  elmo-e  & 98.14 (0.72) & 98.17 (0.72) & 98.14 (0.69) \\ \hline
  elmo-o  & 96.87 (0.90) & 96.93 (0.92) & 96.91 (0.95) \\ \hline
  char    & 97.17 (0.82) & 97.25 (0.84) & 97.23 (0.79) \\ \hline
  \multicolumn{4}{|c|}{{\it test}} \\  \hline
  bert    & 97.95 (0.35) & 97.94 (0.36) & 97.97 (0.35) \\ \hline
  elmo-e  & 98.29 (0.36) & 98.30 (0.36) & 98.30 (0.36) \\ \hline
  elmo-o  & 97.09 (0.45) & 97.14 (0.43) & 97.16 (0.46) \\ \hline
  char    & 97.23 (0.41) & 97.30 (0.42) & 97.28 (0.45) \\ \hline
\end{tabular}
    \caption{Cross-validated \pos accuracy on PPCEME using no function tags, 10 function tags, or 31 function tags. Standard deviations in parentheses. The rows and columns are analogous to those in Table~\ref{tab:cvscores}.}
    \label{tab:cvposscores}
\end{table}

\begin{table}[t]
\footnotesize
\centering
    \begin{tabular}{|c|c|c|} \hline
    tag    & frequency    & f1 \\ \hline
    N      & 11.46 (1.22) & 97.05 (1.02)  \\ \hline
    P      & 11.38 (0.47) & 99.20 (0.40)  \\ \hline
    PRO    &  7.33 (0.97) & 99.75 (0.17)  \\ \hline
    D      &  7.21 (0.57) & 99.58 (0.24)  \\ \hline
    ,      &  7.03 (0.83) & 99.72 (0.18)  \\ \hline
    .      &  5.27 (0.78) & 99.66 (0.21)  \\ \hline
    CONJ   &  5.08 (0.63) & 99.59 (0.21)  \\ \hline
    ADJ    &  4.05 (0.57) & 95.69 (1.28)  \\ \hline
    NPR    &  3.61 (1.75) & 93.80 (1.74)  \\ \hline
    ADV    &  3.41 (0.37) & 96.75 (1.44)  \\ \hline
    NS     &  3.19 (0.49) & 97.43 (0.82)  \\ \hline
    VB     &  2.68 (0.52) & 98.36 (1.06)  \\ \hline
    PRO\$  &  2.48 (0.38) & 99.43 (0.60)  \\ \hline
    VBD    &  2.25 (0.83) & 97.49 (1.39)  \\ \hline
    VAN    &  1.74 (0.17) & 95.64 (2.01)  \\ \hline
    VBP    &  1.72 (0.38) & 96.58 (1.21)  \\ \hline
    MD     &  1.61 (0.42) & 99.53 (0.18)  \\ \hline
    Q      &  1.59 (0.30) & 98.58 (1.47)  \\ \hline
    BEP    &  1.59 (0.38) & 99.03 (0.87)  \\ \hline
    TO     &  1.30 (0.37) & 99.45 (0.62)  \\ \hline
    C      &  1.17 (0.28) & 98.73 (0.48)  \\ \hline
    BED    &  1.00 (0.38) & 99.54 (0.65)  \\ \hline
    WPRO   &  0.90 (0.24) & 99.29 (0.37)  \\ \hline
    NUM    &  0.74 (0.40) & 94.42 (5.00)  \\ \hline
    NEG    &  0.73 (0.15) & 99.68 (0.22)  \\ \hline
    VAG    &  0.69 (0.16) & 95.23 (1.34)  \\ \hline
    VBN    &  0.63 (0.13) & 96.52 (1.17)  \\ \hline
    BE     &  0.59 (0.19) & 99.26 (0.23)  \\ \hline
    HVP    &  0.55 (0.12) & 99.21 (1.19)  \\ \hline
    FW     &  0.54 (0.41) & 84.25 (7.89)  \\ \hline
    RP     &  0.51 (0.12) & 95.73 (1.77)  \\ \hline
    ADVR   &  0.43 (0.19) & 96.86 (0.91)  \\ \hline
    VBI    &  0.40 (0.14) & 91.61 (4.34)  \\ \hline
    WADV   &  0.36 (0.10) & 98.01 (1.62)  \\ \hline
\end{tabular}
    \caption{Cross-validated per-tag F1 of the \elmo(\eebo) configuration for the 34 most frequent \pos tags on the \ppceme dev section (using the 31 function tag set). The frequency column indicates  the mean relative frequency of each tag in the dev set. Standard deviations are in parentheses.}
    \label{tab:cvposbreakdown}
\end{table}

While the evalb scores are of more importance, \pos accuracy also matters as the linguistic queries we expect to be run on the eventual parsed \eebo sometimes refer explicitly to \pos tags. Therefore, in this section we consider the \pos results in detail. As is apparent from Table~\ref{tab:cvposscores}, \pos-accuracy is highest for \elmo(\eebo) for all combinations of function tags and dev/test set, followed by \bert, then \elmo(orig); e.g., when using 31 function tags, \elmo(\eebo) outperforms \bert by 0.33\% absolute (t=5.68, p$<$1e-3) on the test set. 

As the distribution of \pos tags is highly unbalanced, we also report F1 by tag for the 34 most frequent \pos tags (i.e., tags with a frequency $\geq$0.30\%) in the \ppceme dev set (Table~\ref{tab:cvposbreakdown}\footnote{For descriptions of \ppceme \pos tags in the table,  see \url{https://www.ling.upenn.edu/hist-corpora/annotation/labels.htm}.}). While there is some variation among the tags, F1 is consistently high (>95\%) with the exception of the imperative (VBI), proper noun (NPR)\footnote{That NPR should achieve the second lowest F1 (93.80\%) might seem surprising. We theorize that this is the result of confusion between proper nouns and common nouns (N) in the parser's \pos predictions.
This almost certainly arises from some known in the issues in the corpus of inconsistent annotation between N and NPR,  which has not been a focus for correction since it has little-to-no relevance to the linguistic queries that the corpus is used for.  However, the N score of 97.05 does lower our overall \pos score since it is the most common tag.}, cardinal number (NUM), and foreign word (FW). The imperative tag is of particular concern for us as we anticipate future users of the parsed \eebo to rely heavily on it for retrieval of imperative clauses. Consequently, we go into rather greater detail for the imperative and provide two typical examples of parser confusion between VBI and VBP (present).
\subsection{Example 1}
{\vbox
{\footnotesize
\begin{verbatim}
(a) (IP-MAT (CONJ &)
            (NP-SBJ *con*) 
            (VBP stryue)
            (IP-INF (TO to)
                   (VB get)
                   (NP-OB1 (D that)))))
(b) (IP-IMP (CONJ &)
            (VBI stryue)
            (IP-INF (TO to)
                   (VB get)
                   (NP-OB1 (D that)))))
\end{verbatim}}}
In this first example, tree (a) is the gold tree, while (b) is the parser output, which has both the wrong \pos tag for {\it stryue} (strive) and the wrong function tag for the parent IP.  {\tt *con*} in the gold tree is an elided subject under conjunction, and in this case refers to a subject in the previous tree. Generating the correct parse requires, as it would for a human, referring to the preceding text.    In isolation, its parse with the VBI is perfectly reasonable, although incorrect in this case.  

Another aspect of this error concerns the common NLP practice, which we follow, of removing  empty categories from the parser training material (Appendix \ref{app:ppcemepre:empty}). In this case, the removal of {\tt *con*} has the consequence that the training data has ordinary (i.e. non-imperative) sentences without an overt subject and a seemingly arbitrary assignment of VBP or VBI to verbs. 

This example therefore shows both the need to refer to text beyond the sentence level, and how the training data is corrupted by the removal of empty categories.

\subsection{Example 2}

{\vbox
{\footnotesize
\begin{verbatim}
(a) (IP-IMP (PP (P For)
                (NP (NS Mice)))
            (PP (ADV+P therefore))
            (VBI lay)
            (NP-OB1 (NP (N Poyson))
                    (CONJP or Oatmeal
                           mixt
                           with 
                           pounded
                           glass)))
(b) (IP-MAT (CONJ for)
            (NP-SBJ (NS Mice))
            (PP (P therefore))
            (VBP lay)
            (NP-OB1 (NP (N Poyson))...))
\end{verbatim}}}

This second example shows the reverse error, in which the gold \pos tag is VBI for {\it lay}, while the parser makes it a VBP.  The reason is that it incorrectly parses {\it for Mice}, and so thinks the sentence has a NP-SBJ, and then reasonably, given that mistake, makes {\tt lay} a VBP, not an imperative.


.

\section{Out-of-vocabulary span labels and resource inefficiency}

As shown in Table \ref{tab:cvscores}, for each configuration (row) there is little effect on the f1 parsing scores among  \ftagsa, \ftagsb, and \ftagsc, a result that is consistent with \citet{gabbard-etal-2006-fully} for the \ptb.   However, the inclusion of the function tags has a significant effect on the label vocabulary, which we discuss before going on to the analysis of the function tag performance in Section \ref{sec:results:ftags}. 

As discussed in Section \ref{sec:model:parser}, the parser uses a classifier to produce the most likely label for a span. The parser determines the label vocabulary at training time, by collecting all the possible nonterminal labels.  There are two complications in this determination of the label vocabulary.

First, for unary branches, the parser combines the different nonterminals into a single label.  For example, for the structure {\tt (NP (CP ...))}, which is a NP with a single child, a CP, the label is NP::CP.  At parse time, if this label is the most likely for a span, then it is split up again into a parent NP and child CP node. 



Second, since function tags are integrated into the label vocabulary, they multiply out over the cases with unary branching.  For example, with the structure {\tt (NP (CP-FRL ...))}, for a NP with a free relative CP child, the value in the label vocabulary will be NP::CP-FRL.  But there can also be instances {\tt (NP-SBJ (CP-FRL ...))}, {\tt (NP-OB1 (CP-FRL ...))}, resulting in the entries NP-SBJ::CP-FRL and NP-OB1::CP-FRL in the label vocabulary.   

\begin{table}
\footnotesize
    \centering
    \begin{tabular}{|l||r|r||r|r||r|r|} \hline
     &
     \multicolumn{2}{|c||}{\ftagsa} &
     \multicolumn{2}{|c||}{\ftagsb} &
     \multicolumn{2}{|c|}{\ftagsc} \\ \hline
    sec & w/  & w/o & w/ & w/o & w/ & w/o \\ \hline
    train   & 330 & 88 & 542 & 124 & 920 & 250 \\ \hline
    dev & 138 & 48 & 218 & 77 & 363 & 157 \\ \hline
    both & 125 & 48 & 196 & 77 & 323 & 154 \\ \hline
    \end{tabular}
    \caption{Label vocabulary size.  The columns ``w/'' and ``w/o'' are for with or without the collapse of unary nodes for determining the label vocabulary.  The label vocabulary used by the model is the ``w/'' value for the train section.  The last row shows the number of labels that are common to the train and dev sections.}
    \label{tab:labelvocab}
\end{table}

Table \ref{tab:labelvocab} shows the impact of these two factors   on the label vocabulary for the first split.  The ``w/o'' column shows the number of different labels without combining labels on unary branches, while the ``w/'' column shows the number of labels with combined unary branches .   The entry for the ``train'' row and ``w/'' column is the size of the label vocabulary actually used by the parser. 

Even without any function tags, the collapsing of unary branches causes the label vocabulary to increase from 88 to 330.  With all 31 function tags, the label vocabulary increase from 250 for single nodes to 920.  This increases resource requirements during training, and we found that the training time could increase as much as 40\% between trains of \ftagsa and \ftagsc.  

Much of the increased resource requirements is not required, though. 
 The ``dev'' row  shows the corresponding numbers for the dev section, and the ``both'' row shows the label values that are in both the train and dev sections.  Even without function tags, only 125 of the labels in the training section occur in the dev section, while the dev section has 13 labels that are not in the training section.  For \ftagsc,  only 323 of the 920 labels in the training section are in the dev section, and the dev section has 40 labels that are not in the training section.
 
 If a label is in the dev section but not in the label vocabulary determined by the train section, it is impossible for the parser to correctly predict spans with that label.  They are in effect out-of-vocabulary span labels. While the parser score is not not damaged more, due to the infrequency of these OOV  labels, this shows the limits of the current approach. We consider these current results to be a baseline for future improvement.

\section{Function tag analysis}
\label{sec:results:ftags}

\begin{table}[t]
\footnotesize
    \centering
\begin{tabular}{|c|c|c|} \hline
config & ftags-10 & ftags-31 \\ \hline
\multicolumn{3}{|c|}{{\it dev}} \\  \hline
bert & $97.79 (0.60)$ & $94.40 (1.42)$ \\ \hline
elmo-e & $97.94 (0.51)$ & $94.90 (1.54)$ \\ \hline
elmo-o & $97.13 (0.66)$ & $93.56 (1.69)$ \\ \hline
char & $97.01 (0.69)$ & $93.44 (1.60)$ \\ \hline
\multicolumn{3}{|c|}{{\it test}} \\  \hline
bert & $97.72 (0.27)$ & $94.95 (0.94)$ \\ \hline
elmo-e & $97.90 (0.28)$ & $95.55 (0.87)$ \\ \hline
elmo-o & $96.97 (0.28)$ & $94.24 (0.81)$ \\ \hline
char & $96.88 (0.35)$ & $94.22 (0.82)$ \\ \hline
\end{tabular}
    \caption{Function tag scores for the different configurations, showing the mean and standard deviation over the 8 splits.  The function tag score uses a comparison of the function tags for brackets that count as a match for \evalb.  There is no function tag score for \ftagsa since there are no function tags in that version of the training and evaluation data.}
    \label{tab:cvftagscores}
\end{table}

\begin{table}[t]
\footnotesize
\centering
    \begin{tabular}{|r|l|r|r|} \hline
\multicolumn{1}{|c|}{tag} & description & \multicolumn{1}{|c|}{frequency} & \multicolumn{1}{|c|}{f1} \\ \hline
\multicolumn{2}{|c|}{\textbf{Syntactic}} & 38.29 ( 2.82) & 97.44 \ ( 0.78)  \\ \hline
*SBJ & subject & 22.56 ( 1.35) & 98.63 \ ( 0.55)  \\ \hline
*OB1 & accusative & 11.26 ( 0.73) & 96.38 \ ( 0.97)  \\ \hline
*OB2 & dative &  1.23 ( 0.36) & 93.64 \ ( 1.44)  \\ \hline
*VOC & vocative &  1.05 ( 0.55) & 95.26 \ ( 1.62)  \\ \hline
MSR & measure &  1.03 ( 0.29) & 92.92 \ ( 1.71)  \\ \hline
POS & possessive &  0.94 ( 0.37) & 97.77 \ ( 0.99)  \\ \hline
SPR & sec pred &  0.21 ( 0.07) & 77.95 \ ( 5.11)  \\ \hline
\multicolumn{2}{|c|}{\textbf{Semantic}} &  8.41 ( 1.76) & 95.60 \ ( 0.55)  \\ \hline
TMP & temporal &  3.71 ( 1.88) & 94.63 \ ( 1.66)  \\ \hline
ADV & adverbial &  3.27 ( 0.61) & 97.09 \ ( 0.42)  \\ \hline
LOC & locative &  0.88 ( 0.32) & 93.53 \ ( 2.52)  \\ \hline
DIR & directional &  0.55 ( 0.14) & 93.34 \ ( 2.01)  \\ \hline
\multicolumn{2}{|c|}{\textbf{CP only}} &  8.09 ( 0.95) & 94.61 \ ( 1.15)  \\ \hline
REL & rel clause &  3.02 ( 0.40) & 95.08 \ ( 1.47)  \\ \hline
THT & THAT clause &  2.26 ( 0.70) & 97.26 \ ( 1.23)  \\ \hline
*QUE & question &  1.51 ( 0.47) & 96.31 \ ( 0.86)  \\ \hline
CAR & clause-adj &  0.50 ( 0.15) & 77.31 \ ( 4.11)  \\ \hline
CMP & comparative &  0.42 ( 0.17) & 94.62 \ ( 2.51)  \\ \hline
FRL & free relative &  0.30 ( 0.11) & 85.37 \ ( 3.55)  \\ \hline
EOP & empty op &  0.08 ( 0.03) & 94.47 \ ( 5.01)  \\ \hline
\multicolumn{2}{|c|}{\textbf{IP only}} &  7.81 ( 1.62) & 95.79 \ ( 1.73)  \\ \hline
*INF & infinitival &  3.75 ( 1.12) & 98.81 \ ( 0.47)  \\ \hline
PPL & participial &  1.57 ( 0.43) & 97.96 \ ( 1.18)  \\ \hline
*IMP & imperative &  1.14 ( 0.37) & 92.85 \ ( 3.26)  \\ \hline
SMC & small clause &  0.69 ( 0.11) & 97.00 \ ( 1.63)  \\ \hline
PRP & purpose &  0.40 ( 0.10) & 73.92 \ ( 4.10)  \\ \hline
ABS & absolute &  0.26 ( 0.13) & 81.34 \ ( 8.46)  \\ \hline
\multicolumn{2}{|c|}{\textbf{CP or IP}} & 34.84 ( 4.32) & 92.41 \ ( 4.20)  \\ \hline
*MAT & matrix & 14.52 ( 3.27) & 98.75 \ ( 0.39)  \\ \hline
*SUB & subordinate & 13.28 ( 1.71) & 99.01 \ ( 0.17)  \\ \hline
SPE & direct speech &  6.79 ( 5.15) & 46.09 (24.96)  \\ \hline
DEG & degree &  0.25 ( 0.12) & 88.47 \ ( 3.07)  \\ \hline
\multicolumn{2}{|c|}{\textbf{Misc}} &  2.56 ( 0.34) & 85.54 \ ( 2.55)  \\ \hline
*PRN & parenthetical &  2.04 ( 0.48) & 89.25 \ ( 1.75)  \\ \hline
RSP & resumptive &  0.29 ( 0.11) & 62.23 (10.23)  \\ \hline
LFD & left-disl &  0.23 ( 0.09) & 75.83 \ ( 7.20)  \\ \hline
\multicolumn{2}{|c|}{\textbf{Total}} & 100.00 ( 0.00) & 94.90 \ ( 1.54)  \\ \hline
\end{tabular}

    \caption{Function tag breakdown.  The third column shows the frequency of the tag in the dev section, while the fourth column shows its f1 score. The tags are organized into six groups, with the combined frequency and f1 score for the tags in each group.}
    \label{tab:cvftagbreakdown}
\end{table}

Table \ref{tab:cvftagscores} shows the score for the function tags over the different configurations, with either 10 or 31 function tags.   The overall score for the function tags drops drastically between \ftagsb and \ftagsc. For example, for \elmo \eebo  it falls from a mean of 97.94  with \ftagsb to a mean of 94.90 for \ftagsc.

To explore these results in more detail, we pick one cell in the table -  \elmo \eebo with 31 tags, and look at the individual scores for the 31 function tags that make up the 94.90 result. This is shown in Table \ref{tab:cvftagbreakdown}. The tags are organized into  6 groups for convenience of discussion.   The ``syntactic'' and ``semantic'' tags are roughly similar to those groups for the \ptb, as presented in \citet{gabbard-etal-2006-fully}.  The other groups include tags that are very different than in the \ptb, as mentioned in Section \ref{sec:ppceme:ps:ftags}, such as the tags restricted to CP nodes and indicating properties of the CP clause.  The asterisks in Table \ref{tab:cvftagscores} indicate the ten tags included in ftags-10. The third column shows the frequency of each tag or group of tags among all the function tags being scored in the dev section, and the f1 column shows  the score for the tag or group of tags. 

Examining the results for the tags included in \ftagsc but not in \ftagsb, the main cause for the drop in score is the SPE tag for direct speech, which is one of the most common tags, with an average 6.79\% frequency, but with an average score of 46.09 f1. The \ppceme lacks the consistent clues for direct speech (such as quotation marks) that might be available in newswire text that adhered to strict guidelines.  

For example, the sentence from the dev section for the first split {\tt and he shall be welcome} has an IP-MAT-SPE (matrix IP, direct speech) as the root of the entire tree in gold annotation, while in the parser output it is IP-MAT.  The parser (or for that matter, a person) cannot know that the sentence is direct speech without looking at the wider context, which is a conversation spanning a few sentences.  The underlying problem here is the same as with the first example of a \pos error in Section \ref{sec:posresults}.  Future work to recover the direct speech information will need to examine discourse contexts beyond just single sentences in isolation, as the parser does now. 

The other scores show some variance in accuracy, although some of the most common tags (SBJ, SUB, MAT) have the highest scores.  Looking at the tags as groups, there are some  drops in scores among some of the tags that classify CP or IP constituents (e.g. purpose clauses). These require more analysis, although in general the answer to the question of whether the function tag recovery is satisfactory can only be answered by determining if they are accurate enough to ensure the accuracy of linguistic searches on automatically parsed material, part of the larger framework of this work. 

\section{Conclusion and Future Work}
\label{sec:conclusionfuture}

In this work we have presented the first parsing results on the \ppceme, with a focus on challenges posed by its extensive set of function tags.  Adapting an earlier approach to function tag recovery works surprisingly well overall, while we point out some challenges for future work concerning improvements in accuracy and resource efficiency.   

We have also demonstrated the continued importance of in-domain pretraining, as the parser configuration using \elmo trained on in-domain \eebo outperforms the configuration using \bert trained on modern English.  We performed cross-validation to ensure that the results were robust and not an accident of a particular split.  

Future work will proceed along a number of lines.  Regarding the function tags, an obvious approach is to integrate a separate function tag classifier into the parser separate from one for the bare nonterminals and to combine them to produce the full nonterminal labels.  We will also evaluate the function tags by their utility for linguistic searches on automatically-parsed material, as mentioned at the end of Section \ref{sec:results:ftags}. 


\bibliography{ppcbib}
\bibliographystyle{acl_natbib}

\clearpage
\setcounter{page}{1}
\appendix
\section{PPCEME preprocessing}
\label{app:ppcemepre}

\subsection{Metadata}
\label{app:ppcemepre:metadata}

In addition to the changes described in Section \ref{sec:ppceme:ps:ftags}, we removed the metadata under {\tt CODE}, {\tt META}, and {\tt REF} nodes. In cases where {\tt CODE} dominated a leaf, removing the leaf resulted in an ill-formed tree. The 267 trees in question were removed, as were 576 trees rooted in {\tt META} (usually stage directions for a play) and 9 trees containing {\tt BREAK}.

In addition, before carrying out the above modifications, we changed all instances of \verb+(CODE <paren>)+ and \verb+CODE <$$paren>)+ to \verb+(OPAREN -LRB-)+ and \verb+(CPAREN -RRB-)+, respectively. We did this in order to retain the parentheses that otherwise, being daughters of {\tt CODE}, would have been deleted. 

Our counts of number of words and sentences differ slightly from  \citet{yang-eisenstein-2016-part}.  This is probably related to small differences of preparation of the type just discussed.

\subsection{PPCEME Part-of-Speech Modifications}
\label{app:ppcemepre:pos}

In addition to the changes described in the main text, we changed the tag {\tt MD0} to {\tt MD}. {\tt MD0} is an untensed modal, as in {\tt he will can} or {\tt to can do something}. There are only 4 cases, as this is an option that had mostly died out by the time of Early Modern English.


%
There are also cases where words that are ordinarily spelled as a single orthographic token are sometimes split into several tokens. \ppceme represents the former case with a single \pos tag and the latter as a constituent whose non-terminal is the \pos tag, with the words given numbered segmented \pos tags  - for example, {\tt (ADJ alone)} vs. {\tt (ADJ (ADJ21 a) (ADJ22 lone))}. We modified all such tags by removing the numbers, and appending {\tt \_NT} to the nonterminals, in order to more clearly distinguish between \pos tags and nonterminals.  In this example, the resulting structure would be  {\tt (ADJ\_NT (ADJ a) (ADJ lone))}.

\subsection{Function Tags}
\label{app:ppcemepre:ftags}

We exclude certain tags that occur very rarely in \ppceme (CLF, COM, TMC, RFL, ADT, EXL, YYY, ELAB, XXX, TAG, and TPC). Table \ref{app:corpus:ppceme:ftagstab} shows the frequency for each of the remaining 31 tags in the entire corpus for nonterminals with a non-empty yield.  For convenience, the tags are organized into six groups. The syntactic and semantic groups are roughly similar to those groups for the \ptb, as presented in \citet{gabbard-etal-2006-fully}. The other groups include tags that differ significantly from those in the \ptb, as noted in Section \ref{sec:ppceme:ps:ftags}. For the full set of \ppceme function tags, see \url{https://www.ling.upenn.edu/hist-corpora/annotation/labels.htm}.

\begin{table}[t]
\footnotesize
\centering
    \begin{tabular}{|r|l|r|} \hline
Tag & Description & Frequency \\ \hline
\multicolumn{2}{|c}{\textbf{Syntactic}} & \multicolumn{1}{|r|}{\textbf{37.23}}  \\ \hline
SBJ & subject & 21.00  \\ \hline
OB1 & direct object & 11.96  \\ \hline
OB2 & indirect object & 1.20  \\ \hline
SPR & secondary predicate & 0.28  \\ \hline
MSR & measure & 1.17  \\ \hline
POS & possessive & 0.86  \\ \hline
VOC & vocative & 0.77  \\ \hline
\multicolumn{2}{|c}{\textbf{Semantic}} & \multicolumn{1}{|r|}{\textbf{7.93}}  \\ \hline
DIR & directional & 0.50  \\ \hline
LOC & locative & 0.84  \\ \hline
TMP & temporal & 3.09  \\ \hline
ADV & adverbial & 3.50  \\ \hline
\multicolumn{2}{|c}{\textbf{CP only}} & \multicolumn{1}{|r|}{\textbf{8.83}}  \\ \hline
CAR & clause-adjoined & 0.55  \\ \hline
REL & relative clause & 3.36  \\ \hline
THT & THAT clause & 2.52  \\ \hline
CMP & comparative & 0.53  \\ \hline
QUE & question & 1.35  \\ \hline
FRL & free relative & 0.33  \\ \hline
EOP & empty operator & 0.19  \\ \hline
\multicolumn{2}{|c}{\textbf{IP only}} & \multicolumn{1}{|r|}{\textbf{9.67}}  \\ \hline
INF & infinitive & 4.59  \\ \hline
PPL & participial & 2.18  \\ \hline
IMP & imperative & 1.12  \\ \hline
SMC & small clause & 0.90  \\ \hline
PRP & purpose & 0.46  \\ \hline
ABS & absolute & 0.42  \\ \hline
\multicolumn{2}{|c}{\textbf{CP or IP}} & \multicolumn{1}{|r|}{\textbf{33.10}}  \\ \hline
SUB & subordinate & 14.52  \\ \hline
MAT & matrix & 12.66  \\ \hline
SPE & direct speech & 5.64  \\ \hline
DEG & degree & 0.28  \\ \hline
\multicolumn{2}{|c}{\textbf{Miscellaneous}} & \multicolumn{1}{|r|}{\textbf{3.23}}  \\ \hline
PRN & parenthetical & 2.60  \\ \hline
RSP & resumptive & 0.33  \\ \hline
LFD & left-dislocated & 0.30  \\ \hline
\end{tabular}
    \caption{Relative frequencies of the 31 retained function tags in \ppceme. The tags are organized into 6 groups, with combined frequency by group in boldface.}
    \label{app:corpus:ppceme:ftagstab}
\end{table}

\subsection{Empty categories}
\label{app:ppcemepre:empty}
\ppceme indicates discontinuous dependencies by means of empty categories that are coindexed with a displaced constituent.
Following common NLP practice, we remove both the empty categories and the co-indexing from the parser training material, and thus from the parser output.   This simplifies the parsing model, and for present purposes,  the absence of empty categories is irrelevant.
However, if we wish to include linguistic queries in future work that make reference to empty categories, as is necessary in the general case, the parsing model will need to be augmented appropriately.

\section{Cross-validation Splits}
\label{app:splits}

\begin{table*}[t]
        \centering
        \begin{tabular}{|l|rr|rr|rr|} \hline
                section   & \multicolumn{2}{|c|}{\# files} & \multicolumn{2}{|c|}{\# tokens} & \multicolumn{2}{|c|}{\% of split} \\ \hline
                train & 205.88 & (13.34) &  1743211.25 & (10441.53) &  89.65 & (0.54) \\ \hline
                dev & 12.50 & (7.15) &  101000.12 & (4081.82) &  5.19 & (0.21) \\ \hline
                test & 13.62 & (7.91) &  100268.62 & (7832.66) &  5.16 & (0.40) \\ \hline
                OVERALL & 232 & (0.00) &  1944480 & (0.00) &  100 & (0.00) \\ \hline
        \end{tabular}
        \caption{Mean number of files and tokens for train/dev/test sections across the 8 cross-validation splits (standard deviations are presented in parentheses). The percentage of tokens in each section is also presented (in the {\bf \% of split} column).}
        \label{tab:splits1}
\end{table*}

Table \ref{tab:splits1} summarizes the composition of the train/dev/test sections across the cross-validation 8 splits; specifically, the total number of documents, the total number of tokens, and the percentage of total tokens in each section. Since the partitioning process is performed at the level of \ppceme source files, and these files differ substantially in size, there is some variation in these numbers across the splits. For this reason, we report standard deviations as well as means.  The final row (``OVERALL'')  depicts numbers for a complete split (i.e., the train/dev/test sections combined); as this is constant across each split, the entries of this row have a standard deviation of zero.  As can be seen, overall the splits have kept to target 90-5-5 breakdown; e.g., the train section on average comprises 89.65\% of the total tokens with a standard deviation of 0.54\%.

As mentioned in the main text, the corpus consists of text from three main time periods (e1, e2, e3)\footnote{For details regarding the \ppceme time periods (e1, e2 and e3) see \url{https://www.ling.upenn.edu/hist-corpora/PPCEME-RELEASE-3/description.html}} and we aimed to balance the time periods equally within each split, to the extent possible given that we treated the files as atomic units. Table \ref{tab:splits2} shows the breakdown by period. Similar to Table~\ref{tab:splits1}, mean/standard deviation for total number of documents/tokens are presented for each time period in each section. Additionally, for each time period, it reports the mean percentage of each split (in tokens) from each time period. The marginals provide numbers combining across time periods (the ``ALL PERIODS'' row) and sections (the ``ENTIRE SPLIT'' column). For example, the training section contains on average 1,743,211.25 tokens, with on average 32.85\% coming from time period e1, 36.61\% from e2, and 30.53\% from e3.

\begin{table*}[t]
        \centering
        {\tiny
        \begin{tabular}{|l||r|r|r||r|r|r||r|r|r||r|r|r||} \hline
                 & \multicolumn{3}{|c||}{train section} & \multicolumn{3}{|c||}{dev section} & \multicolumn{3}{|c||}{test section} & \multicolumn{3}{|c||}{ENTIRE SPLIT} \\ \cline{2-13}
                 period   & \# files  & \# tokens   & \% train   & \# files   & \# tokens   & \% dev   & \# files   & \# tokens   & \% test   & \# files   & \# tokens   & \% split \\ \hline
                 e1       & 72.88     & 572672.62   & 32.85      & 4.25       & 33178.50    & 32.98    & 4.88       & 31369.88    & 31.50     & 82         & 637221      & 32.77 \\
                          & (6.51)    & (11974.31)  & (0.79)     & (3.01)     & (7078.50)   & (7.36)   & (4.55)     & (8193.65)   & (8.67)    & (0.00)     & (0.00)      & (0.00) \\ \hline
                 e2       & 66.00     & 638269.88   & 36.61      & 4.00       & 34844.62    & 34.40    & 4.00       & 35186.50    & 34.89     & 74         & 708301      & 36.43 \\
                          & (4.38)    & (13490.18)  & (0.60)     & (2.51)     & (6382.81)   & (5.41)   & (2.14)     & (7767.44)   & (6.18)    & (0.00)     & (0.00)      & (0.00) \\ \hline
                 e3       & 67.00     & 532268.75   & 30.53      & 4.25       & 32977.00    & 32.63    & 4.75       & 33712.25    & 33.60     & 76         & 598958      & 30.80 \\
                          & (5.18)    & (7066.41)   & (0.35)     & (3.96)     & (5211.71)   & (4.65)   & (3.45)     & (5592.81)   & (4.70)    & (0.00)     & (0.00)      & (0.00) \\ \hline
                 ALL      & 205.88    & 1743211.25  & 100        & 12.50      & 101000.12   & 100      & 13.62      & 100268.62   & 100       & 232        & 1944480     & 100 \\
                 PERIODS  & (13.34)   & (10441.53)  & (0.00)     & (7.15)     & (4081.82)   & (0.00)   & (7.91)     & (7832.66)   & (0.00)    & (0.00)     & (0.00)      & (0.00) \\ \hline
        \end{tabular}}
        \caption{Mean number of files and tokens for train/dev/test sections within each of three time periods (e1, e2, and e3) across the 8 cross-validation splits. The {\bf \% train/dev/test} columns indicate the \% of total train/dev/test tokens for each time period. Standard deviations are presented in parentheses.}
        \label{tab:splits2}
\end{table*}


\section{ELMo Pretraining on EEBO}
\label{app:elmoeebo}

\subsection{Text Extraction}
\label{app:elmoeebo:extract}
\eebo's XML files contain a great deal of metadata and markup in addition to the source text. For each file, we extracted the core source information (title, author, date) and kept the text within \verb+<P>+ tags, which gives at least a rough sense of the document divisions. 

Following \citet[pp. 105-6]{ecay}, we excluded some metadata and other material embedded in the text. Information under {\tt NOTE}, {\tt SPEAKER}, and {\tt GAP} elements were eliminated, as was information under the {\tt L} (``line of verse'') element, which was considered irrelevant for the linguistic searches envisioned for the final resource\footnote{In future work, we will likely revise this to keep the text but with some meta-tags to indicate its origin. }. We also adopting his handling of {\tt GAP} tags, which indicate the locations of OCR errors, which consists of mapping OCR errors to word-internal bullet characters - e.g.,  \verb+Eccl•siasticall+. 
%

\subsection{Normalization}
\label{app:elmoeebo:norm}
The extracted text underwent Unicode normalization to NFC form in order to eliminate spurious surface differences between tokens. The resulting text contained 642 unique characters, 381 of which occurred fewer than 200 times. Manual inspection of these uncommon characters revealed that while some of these made sense in context (e.g., within sections of Greek or Latin text), many seemed to be spurious characters due to OCR errors (e.g., {\sc white rectangle 0x25ad}). Consequently, we elected to filter out all sentences containing characters occurring fewer than 200 times.  This eliminated 4139 lines, with 9,341,966 remaining for training (consisting of 1,168,749,620 tokens).

\subsection{Tokenization}
\label{app:elmoeebo:token}
After normalizing the extracted text into Unicode NFC form in order to eliminate spurious surface differences between tokens, we tokenized the \eebo text in accordance with \ppceme's tokenization guidelines as best we could:
    \begin{enumerate}
        \item Possessive morphemes are not separated from their host (e.g., {\tt Queen{\vtick}s}) (unlike in \ptb).
        \item Punctuation is separated except in the case of abbreviations (e.g., {\tt Mr.}), token-internal hyphens  (e.g., {\tt Fitz-Morris}), or  certain special cases (e.g., {\tt \&c}).
        \item Roman numerals can include leading, internal, or trailing periods 
        (e.g., {\tt .xiiii.C.}).
    \end{enumerate}
    
\ppceme tokenization is straightforward in principle, but the non-standardized nature of the historical material raises various difficulties.  For instance, it is easy to tell that the elided article {\tt th{\vtick}} should be split off (e.g., {\tt th{\vtick}exchaung} is tokenized as {\tt th{\vtick} exchaung}).  But when the apostrophe is missing, the status of {\tt th} is unclear (e.g., {\tt thafternoone} is tokenized as {\tt th afternoone}, but {\tt thynkyth} remains a single token). Another example of pervasive ambiguity is {\tt its} and {\tt it{\vtick}s}; in \ppceme, these forms were  tokenized manually as one token or two, depending on whether the spelling represents the possessive form of the pronoun {\it it} or the contracted form of {\it it is}.  Since \eebo's size rules out manual processing, we resolved such ambiguities by defaulting to the more common case. In the above examples, this resulted in  splitting the variants with apostrophes and not splitting the ones without.\footnote{Future work could consider a joint tokenization-\pos-tagging model.}



\section{Parser}
\label{app:model:parser}

Table \ref{tab:hyperparameters} shows the hyperparameter settings used in the Berkeley Neural Parser. These are all the default settings for these parameters.  We added a parameter {\tt max\_epochs}, used to set the maximum number of epochs.    For the cross-validation training reported, we set {\tt max\_epochs}$=50$.

\begin{table}[t]
\footnotesize
    \centering
    \begin{tabular}{|l|r|} \hline
        hyperparameter & value \\ \hline
attention\_dropout & 0.2    \\ \hline
batch\_size & 32    \\ \hline
char\_lstm\_input\_dropout & 0.2    \\ \hline
checks\_per\_epoch & 4    \\ \hline
clip\_grad\_norm & 0.0    \\ \hline
d\_char\_emb & 64    \\ \hline
d\_ff & 2048    \\ \hline
d\_kv & 64    \\ \hline
d\_label\_hidden & 256    \\ \hline
d\_model & 1024    \\ \hline
d\_tag\_hidden & 256    \\ \hline
elmo\_dropout & 0.5    \\ \hline
encoder\_max\_len & 512    \\ \hline
force\_root\_constituent & 'auto'    \\ \hline
learning\_rate & 5e-05    \\ \hline
learning\_rate\_warmup\_steps & 160    \\ \hline
max\_consecutive\_decays & 3    \\ \hline
max\_len\_dev & 0    \\ \hline
max\_len\_train & 0    \\ \hline
morpho\_emb\_dropout & 0.2    \\ \hline
num\_heads & 8    \\ \hline
num\_layers & 8    \\ \hline
predict\_tags & True    \\ \hline
relu\_dropout & 0.1    \\ \hline
residual\_dropout & 0.2    \\ \hline
step\_decay\_factor & 0.5    \\ \hline
step\_decay\_patience & 5    \\ \hline
tag\_loss\_scale & 5.0    \\ \hline
max\_epochs & 50    \\ \hline
\end{tabular}
    \caption{Hyperparameters Used with the Berkeley Neural Parser.}
    \label{tab:hyperparameters}
\end{table}

\section{Pretraining comparison experiments}
\label{app:pretraining}
Table~\ref{tab:cvscores2} expands Table~\ref{tab:cvscores} to include recall and precision in addition to F1.

\begin{table}[t]
\footnotesize
    \centering
\begin{tabular}{|r|c|c|c|} \hline
config & ftags-0 & ftags-10 & ftags-31 \\ \hline \hline
\multicolumn{4}{|c|}{{\it dev}} \\  \hline\hline
bert (f1) & $90.37 (1.95)$ & $90.37 (1.95)$ & $90.37 (1.98)$ \\ \hline
(rec) & $90.09 (2.05)$ & $90.02 (2.06)$ & $90.00 (2.08)$ \\ \hline
(prec) & $90.65 (1.84)$ & $90.72 (1.84)$ & $90.74 (1.87)$ \\ \hline
\hline
elmo-e (f1) & $90.92 (1.86)$ & $90.95 (1.79)$ & $90.89 (1.83)$ \\ \hline
(rec) & $90.57 (1.96)$ & $90.64 (1.91)$ & $90.49 (1.93)$ \\ \hline
(prec) & $91.27 (1.76)$ & $91.26 (1.69)$ & $91.30 (1.74)$ \\ \hline
\hline
elmo-o (f1) & $88.06 (2.23)$ & $88.13 (2.25)$ & $88.09 (2.22)$ \\ \hline
(rec) & $87.47 (2.35)$ & $87.51 (2.45)$ & $87.38 (2.37)$ \\ \hline
(prec) & $88.66 (2.12)$ & $88.76 (2.06)$ & $88.81 (2.07)$ \\ \hline
\hline
char (f1) & $87.28 (2.34)$ & $87.43 (2.33)$ & $87.33 (2.33)$ \\ \hline
(rec) & $86.80 (2.45)$ & $86.86 (2.49)$ & $86.70 (2.46)$ \\ \hline
(prec) & $87.76 (2.24)$ & $88.02 (2.18)$ & $87.98 (2.21)$ \\ \hline
\hline
\multicolumn{4}{|c|}{{\it test}} \\  \hline\hline
bert (f1) & $89.81 (0.76)$ & $89.83 (0.80)$ & $89.86 (0.77)$ \\ \hline
(rec) & $89.51 (0.85)$ & $89.46 (0.87)$ & $89.44 (0.86)$ \\ \hline
(prec) & $90.11 (0.67)$ & $90.20 (0.74)$ & $90.29 (0.68)$ \\ \hline
\hline
elmo-e (f1) & $90.56 (0.66)$ & $90.62 (0.71)$ & $90.53 (0.69)$ \\ \hline
(rec) & $90.23 (0.73)$ & $90.31 (0.80)$ & $90.13 (0.76)$ \\ \hline
(prec) & $90.88 (0.60)$ & $90.93 (0.64)$ & $90.92 (0.62)$ \\ \hline
\hline
elmo-o (f1) & $87.31 (0.85)$ & $87.50 (0.84)$ & $87.38 (0.81)$ \\ \hline
(rec) & $86.71 (0.92)$ & $86.92 (0.92)$ & $86.69 (0.90)$ \\ \hline
(prec) & $87.91 (0.79)$ & $88.10 (0.80)$ & $88.08 (0.73)$ \\ \hline
\hline
char (f1) & $86.29 (0.93)$ & $86.50 (0.96)$ & $86.45 (0.90)$ \\ \hline
(rec) & $85.83 (1.11)$ & $85.89 (1.06)$ & $85.79 (1.05)$ \\ \hline
(prec) & $86.77 (0.76)$ & $87.12 (0.88)$ & $87.12 (0.78)$ \\ \hline
\end{tabular}
    \caption{Cross-validation mean and standard deviation f1, recall, and precision parsing scores for each of the three versions of \ppceme, with no function tags, 10 function tags, or 31 function tags. ``bert'' is bert-base, ``elmo-e'' is elmo trained on \eebo, ``elmo-o'' is elmo with the original embeddings, and ``char'' is char-lstm. The f1 scores are the same as in Table \ref{tab:cvscores}.}
    \label{tab:cvscores2}
\end{table}

\section{Part-of-speech Results}
\label{app:posresults}

Table \ref{tab:cvposbreakdownall} shows the complete breakdown of the 98.14 \pos score discussed in Section \ref{sec:posresults}.

\begin{table*}[t]
\footnotesize
\centering
\begin{tabular}{|l|c|c||l|c|c|} \hline
\multicolumn{1}{|c|}{tag} & \multicolumn{1}{|c|}{frequency} & \multicolumn{1}{|c|}{f1} & \multicolumn{1}{|c|}{tag} & \multicolumn{1}{|c|}{frequency} & \multicolumn{1}{|c|}{f1} \\ \hline
TOT & 100.00 ( 0.00) & 98.14 ( 0.69)  & DOD &  0.19 ( 0.10) & 99.49 ( 0.80) \\ \hline
N & 11.46 ( 1.22) & 97.05 ( 1.02)  & HV &  0.19 ( 0.07) & 99.20 ( 0.85) \\ \hline
P & 11.38 ( 0.47) & 99.20 ( 0.40)  & FP &  0.18 ( 0.05) & 95.50 ( 2.08) \\ \hline
PRO &  7.33 ( 0.97) & 99.75 ( 0.17)  & QR &  0.17 ( 0.06) & 98.94 ( 0.61) \\ \hline
D &  7.21 ( 0.57) & 99.58 ( 0.24)  & OPAREN &  0.16 ( 0.10) & 100.00 ( 0.00) \\ \hline
, &  7.03 ( 0.83) & 99.72 ( 0.18)  & CPAREN &  0.16 ( 0.10) & 99.96 ( 0.11) \\ \hline
. &  5.27 ( 0.78) & 99.66 ( 0.21)  & SUCH &  0.15 ( 0.07) & 99.75 ( 0.60) \\ \hline
CONJ &  5.08 ( 0.63) & 99.59 ( 0.21)  & ADJR &  0.15 ( 0.07) & 94.04 ( 1.26) \\ \hline
ADJ &  4.05 ( 0.57) & 95.69 ( 1.28)  & ALSO &  0.15 ( 0.06) & 99.73 ( 0.32) \\ \hline
NPR &  3.61 ( 1.75) & 93.80 ( 1.74)  & NPR\$ &  0.14 ( 0.10) & 85.86 ( 6.52) \\ \hline
ADV &  3.41 ( 0.37) & 96.75 ( 1.44)  & NPRS &  0.14 ( 0.10) & 79.03 (12.08) \\ \hline
NS &  3.19 ( 0.49) & 97.43 ( 0.82)  & ADJS &  0.13 ( 0.08) & 96.11 ( 1.85) \\ \hline
VB &  2.68 ( 0.52) & 98.36 ( 1.06)  & WD &  0.13 ( 0.04) & 97.00 ( 1.80) \\ \hline
PRO\$ &  2.48 ( 0.38) & 99.43 ( 0.60)  & BAG &  0.13 ( 0.03) & 99.51 ( 0.34) \\ \hline
VBD &  2.25 ( 0.83) & 97.49 ( 1.39)  & QS &  0.12 ( 0.05) & 97.64 ( 3.28) \\ \hline
VAN &  1.74 ( 0.17) & 95.64 ( 2.01)  & BEN &  0.10 ( 0.04) & 99.46 ( 0.76) \\ \hline
VBP &  1.72 ( 0.38) & 96.58 ( 1.21)  & DO &  0.09 ( 0.02) & 99.45 ( 0.56) \\ \hline
MD &  1.61 ( 0.42) & 99.53 ( 0.18)  & OTHERS &  0.06 ( 0.01) & 89.90 ( 7.50) \\ \hline
Q &  1.59 ( 0.30) & 98.58 ( 1.47)  & DAN &  0.05 ( 0.02) & 96.58 ( 5.73) \\ \hline
BEP &  1.59 ( 0.38) & 99.03 ( 0.87)  & WPRO\$ &  0.04 ( 0.02) & 99.66 ( 0.73) \\ \hline
TO &  1.30 ( 0.37) & 99.45 ( 0.62)  & WQ &  0.03 ( 0.01) & 91.24 ( 6.50) \\ \hline
C &  1.17 ( 0.28) & 98.73 ( 0.48)  & WARD &  0.03 ( 0.02) & 87.68 ( 4.62) \\ \hline
BED &  1.00 ( 0.38) & 99.54 ( 0.65)  & HAG &  0.03 ( 0.01) & 99.17 ( 1.19) \\ \hline
WPRO &  0.90 ( 0.24) & 99.29 ( 0.37)  & FOR &  0.03 ( 0.01) & 94.21 ( 2.57) \\ \hline
NUM &  0.74 ( 0.40) & 94.42 ( 5.00)  & DON &  0.03 ( 0.01) & 98.58 ( 1.45) \\ \hline
NEG &  0.73 ( 0.15) & 99.68 ( 0.22)  & NS\$ &  0.03 ( 0.02) & 63.70 (15.64) \\ \hline
VAG &  0.69 ( 0.16) & 95.23 ( 1.34)  & \$ &  0.03 ( 0.03) & 76.39 (20.36) \\ \hline
VBN &  0.63 ( 0.13) & 96.52 ( 1.17)  & ADVS &  0.02 ( 0.01) & 90.43 ( 9.80) \\ \hline
BE &  0.59 ( 0.19) & 99.26 ( 0.23)  & X &  0.02 ( 0.03) &  5.28 ( 9.01) \\ \hline
HVP &  0.55 ( 0.12) & 99.21 ( 1.19)  & HVN &  0.01 ( 0.01) & 94.67 ( 3.32) \\ \hline
FW &  0.54 ( 0.41) & 84.25 ( 7.89)  & DOI &  0.01 ( 0.01) & 88.48 (11.01) \\ \hline
RP &  0.51 ( 0.12) & 95.73 ( 1.77)  & BEI &  0.01 ( 0.01) & 89.88 ( 8.78) \\ \hline
ADVR &  0.43 ( 0.19) & 96.86 ( 0.91)  & ' &  0.01 ( 0.01) & 72.75 (36.30) \\ \hline
VBI &  0.40 ( 0.14) & 91.61 ( 4.34)  & ELSE &  0.01 ( 0.01) & 93.14 (11.63) \\ \hline
WADV &  0.36 ( 0.10) & 98.01 ( 1.62)  & HAN &  0.01 ( 0.00) & 94.71 ( 4.21) \\ \hline
HVD &  0.27 ( 0.07) & 99.43 ( 0.60)  & DAG &  0.01 ( 0.00) & 48.18 (21.06) \\ \hline
OTHER &  0.25 ( 0.03) & 96.35 ( 5.76)  & NPRS\$ &  0.01 ( 0.01) & 47.24 (45.31) \\ \hline
N\$ &  0.25 ( 0.10) & 91.11 ( 5.40)  & ONES &  0.01 ( 0.01) & 93.30 ( 9.75) \\ \hline
ONE &  0.25 ( 0.05) & 97.68 ( 2.14)  & HVI &  0.01 ( 0.00) & 87.95 (11.50) \\ \hline
DOP &  0.23 ( 0.09) & 99.17 ( 0.57)  & OTHERS\$ &  0.00 ( 0.00) &  0.00 ( 0.00) \\ \hline
EX &  0.20 ( 0.06) & 96.21 ( 3.65)  & OTHER\$ &  0.00 ( 0.00) & 90.20 (14.48) \\ \hline
" &  0.20 ( 0.32) & 85.67 (37.78)  & ONE\$ &  0.00 ( 0.00) & 61.43 (45.62) \\ \hline
INTJ &  0.20 ( 0.11) & 92.69 ( 4.35)  &  & & \\ \hline
\end{tabular}
\caption{The mean f1 and standard deviation for individual tags making up the dev section (elmo-e, ftags-31) 98.14 score in Table \ref{tab:cvposscores}.}
\label{tab:cvposbreakdownall}
\end{table*}

\end{document}